\documentclass[11pt]{article}
\usepackage{consilr2024}
\usepackage{times}
\usepackage{url}
\usepackage{seqsplit}
\usepackage{amssymb}
\usepackage{amsmath}
\usepackage{latexsym}
\usepackage{authblk}
\usepackage{enumitem}
\usepackage{hyperref}
\usepackage{etoolbox}
\usepackage{graphicx}
\usepackage{subcaption}
\usepackage{float}

\newcommand{\keyword}[1]{
\hspace{0.2cm}%
\fontsize{10}{12}\selectfont%
\textbf{Keywords: } %
}
\title{\textbf{F5-TTS-RO: EXTENDING F5-TTS TO ROMANIAN TTS VIA LIGHTWEIGHT INPUT ADAPTATION}}
\author{RADU GABRIEL CHIVEREANU}
\author{TIBERIU BOROS}

\affil{Research Institute for Artificial Intelligence “Mihai Drăgănescu”\\
\texttt{radu.chivereanu@racai.ro, tibi@racai.ro}}

\date{}

\begin{document}
\maketitle
\begin{abstract} 
This work introduces a lightweight input-level adapter for the F5-TTS model that enables Romanian Language support. To preserve the existing capabilities of the model (voice cloning, English and Chinese support), we keep the original weights frozen, append a sub-network to the model and train it as an extension for the textual embedding matrix of the text encoder. For simplicity, we rely on ConvNeXt module implemented in F5-TTS to also model the co-dependencies between the new character-level embeddings. The module serves as a ``soft'' letter-to-sound layer, converting Romanian text into a continuous representation that the F5-TTS model uses to produce naturally sounding Romanian utterances.

We evaluate the model with a pool of 20 human listeners across three tasks: (a) audio similarity between reference and generated speech, (b) pronunciation and naturalness and (c) Romanian-English code-switching. The results indicate that our approach maintains voice cloning capabilities and enables, to a certain extent, code-switching within the same utterance; however, residual English accent characteristics remain. We open-source our code and provide example audio samples at https://github.com/racai-ro/Ro-F5TTS. 
\end{abstract}


\section{Introduction}
\label{intro}

Text to speech (TTS) technology has advanced rapidly in recent years, using \textbf{a wide variety of architectures based on neural networks}. Beyond text-based conditioning, several approaches incorporate audio-based conditioning to capture and reproduce speaker-specific characteristics in synthesized speech, enabling zero-shot multi-speaker TTS (ZS-TTS) \cite{voice_cloning}  and effective voice cloning with limited access to speaker samples (1-5 seconds of audio).

In this work, \textbf{we focus on extending the F5-TTS model} \cite{f5tts}, an open-source Non-Autoregressive generative framework, to the Romanian language. F5-TTS builds upon E2-TTS \cite{e2tts}, introducing a ConvNeXt module for text encoding and a multimodal DiT module \cite{dit} as backbone. It operates with characters as tokens, padded with filler tokens to match the length of the mel spectrograms.

As a result, the model can be trained using only paired audio–text data, without requiring additional fine-grained annotations. Moreover, F5-TTS demonstrates strong capabilities in speaker identity cloning.

To adapt F5-TTS for the Romanian language, we introduce a lightweight and non-invasive ConvNeXt based adapter that operates only at the input level, leaving the model’s internal layers frozen. The adapter maps Romanian characters into a continuous input space that the model can process, effectively performing a soft letter-to-sound conversion while leveraging and interpolating between the model’s previously learned letter-to-latent representations.

Keeping the core model frozen also \textbf{reduces the risk of overfitting to speakers} in the training set. An additional advantage of this approach is its natural support for code-switching, which is the ability of a model to speak two or more languages in the same utterance, an increasingly important capability given the widespread prevalence of bilingualism \cite{bilingualism}.

\textbf{Our contribution is the following:} We propose a simple and lightweight adapter for F5-TTS that enables Romanian speech synthesis with voice-cloning capabilities. We conduct \textbf{subjective evaluations} of speaker similarity and pronunciation with a panel of 16 native speakers in three distinct tasks. We perform \textbf{objective evaluations} using the Word Error Rate (WER) on transcribed audio and measures of speaker similarity.

\section{Related Work}
State-of-the-art open-source zero-shot text-to-speech (ZS-TTS) systems, such as \cite{xtts,yourtts,f5tts}, currently lack explicit support for Romanian. 
Existing work on ZS-TTS for Romanian typically relies on adapting large pretrained models and architectures to address data scarcity and related challenges. For example, \cite{eff_fastpitch} extended the FastPitch architecture \cite{fastpitch} to Romanian and enabled multi-speaker synthesis by conditioning on speaker embeddings extracted with TitaNet-L \cite{titanet}.  

A central difficulty in adapting pretrained models to new tasks is \textit{catastrophic forgetting}, where fine-tuning degrades previously acquired knowledge \cite{forgetting}. In the context of speech, several strategies have been proposed, including model merging, experience replay, and adapter-based approaches \cite{forgetting_spken_llm,asr_adapters,peft_tts,exp_replay}. Replay buffers are particularly effective, but they depend on storing samples from prior training languages. In contrast, our work focuses on adaptation using only new language data, \textbf{without retaining or revisiting past examples}.  

These adaptation strategies are also relevant in \textit{code-switching} scenarios, where maintaining knowledge of multiple languages is crucial. Current methods typically leverage pretrained multilingual text encoders \cite{codeswitch1} or explicit multi-language training \cite{code_swtich2}. Our approach instead combines the input embedding matrix from previously learned languages with the output of a newly trained input adapter. However, since the model is not explicitly optimized to align embedding spaces, its code-switching ability remains limited.

\section{Proposed Methodology}

Our method involves using a ConvNext module as an input level adapter to align representations with the learned space of embeddings of the model.

\subsection{Dataset}
For our experiments we rely on the SWARA Speech Corpus \cite{swara}, which comprises over 21 hours of high-quality recordings from 17 speakers. This dataset is particularly suitable for demonstrating that our adapter retains speaker cloning capability, as its limited speaker diversity presents a heightened risk of overfitting.

For evaluation, we used examples from the Common Voice 22.0 dataset \cite{common_voice}, the Romanian subset.

\subsection{Model Architecture}
To preserve the full capabilities of F5-TTS, we keep it frozen and train only a lightweight module that maps the Romanian text input into the input space of the model. Let $x = (c_1, c_2, \dots, c_T)$ denote the input Romanian text sequence of length $T$, where $c_i$ is a character, and let $E \in \mathbb{R}^{|V| \times d_e}$ be the trainable embedding layer. Each character is first assigned to a continuous embedding vector $\mathbf{h}_0 = E(x) = \big[E(c_1), E(c_2), \dots, E(c_T)\big] \in \mathbb{R}^{T \times d_e}$. To model dependencies between character embeddings, we apply a ConvNeXt architecture with \textbf{1D convolutions} along the temporal dimension (implementation from the F5-TTS repository), producing contextualized embeddings $\mathbf{h}_\text{ctx} = \text{ConvNeXt}(\mathbf{h}_0) \in \mathbb{R}^{T \times d_h}$. 

Finally, the output is fed into the frozen F5-TTS model to generate speech: $\hat{y} = \text{TTS}(\mathbf{h}_\text{ctx})$. In short, the entire pipeline is compactly represented as $\hat{y} = \text{TTS}(\text{ConvNeXt}(E(x)))$. Figure \ref{fig:logo} displays a visual representation of our approach.

\begin{figure}[H]

\centering
\includegraphics[width=0.9\textwidth]{illustrations/RoF5TTS.png}
  \caption{Overview of our method (right) compared to the original F5-TTS approach (left). We replace the model's initial character level embedding matrix with a new learnable matrix with a ConvNeXt on top.}
  \label{fig:logo}
\end{figure}

\subsubsection{Code Switching}
We also evaluated the model’s ability to generate code-switching examples between Romanian and English. Let $x = (c_1, c_2, \dots, c_T)$ be the input text sequence containing Romanian and English characters, and let $m_R$ and $m_E$ be masks indicating Romanian and English characters, respectively. We use the special token \~\ to signal language switching. The embeddings are computed as follows:

\[
\mathbf{h}_R = \text{ConvNeXt}(E_R(x) \odot m_R), \quad 
\mathbf{h}_E = E_\text{TTS}(x \odot m_E),
\]

where $E_R$ is the trainable embedding module for Romanian characters and $E_\text{TTS}$ is the frozen embedding layer of the TTS model for English characters. The final sequence is obtained by merging the embeddings:

\[
\mathbf{h}_\text{cs} = \mathbf{h}_R + \mathbf{h}_E, \quad \hat{y}_\text{cs} = \text{TTS}(\mathbf{h}_\text{cs}).
\]

This approach is limited, as the language module and the model itself were not explicitly optimized to handle embeddings that represent English and Romanian at the same time. We evaluated the naturalness of the approach in our experiments and found promising results.

\subsection{Training}
The training objective involved applying the original generative flow-matching objective of F5-TTS to the SWARA Corpus.
Hyperparameters are shown in Table \ref{hyperparameters-table}.

\fontsize{10}{12}{
\begin{table}[H]

\begin{center}
\caption{Training hyperparameters}
\label{hyperparameters-table}
\begin{tabular}{|p{6cm}|p{6cm}|}
\hline \bf Hyperparameter & \bf Value \\ \hline
Learning rate & $1 \times 10^{-4}$ \\ \hline
Steps & 40500 \\ \hline
Batch size (audio frames) & 16384 \\ \hline
Warmup updates & 50 \\ \hline
Max samples & 128 \\ 
\hline
\end{tabular}

\end{center}
\end{table}
}

The training was executed with \textbf{1 NVidia A100 GPU}, running for 12 hours.

\section{Experiments}

We conducted a series of experiments to evaluate the performance of our solution. The tests are divided between an subjective evaluation campaign (Section \ref{evalution:subjective}) using volunteers and an objective evaluation metric (Section \ref{evaltion:objective}), namely the automatically computed word error rate (WER) and speaker similarity.

\subsection{Subjective Evaluation}
\label{evalution:subjective}

For subjective evaluation, leveraging a pool of 16 native speakers we defined three tasks: (a) speaker similarity; (b) naturalness and (c) code-switching. 

For comparison, we used MMS-TTS-RON, the Romanian checkpoint from the Massively Multilingual Speech (MMS-TTS) project \cite{mmstts}, along with the base F5-TTS model. MMS-TTS-RON was chosen because it explicitly supports Romanian, being a VITS \cite{vits} model trained on a Romanian corpus.

In the speaker similarity tests, we added samples from original speakers. Thus, our low anchor (a system designed for lower-bound normalization of the results) was the unmodified F5-TTS and for speaker similarity, the original voice was used for high-bound normalization. 

The testing procedure for each task consisted in selecting a list of utterances that were synthesized or extracted from the dataset - for cases where natural speech samples were used. In each evaluation step, the listener would be presented with 3 (or 4) parallel samples, generated by each system (respectively natural). The listener had to evaluate the parallel samples with a score from 0 (Bad) to 100 (Excellent), not knowing the model used for synthesis. 

In what follows, we detail the speaker similarity task in Subsection \ref{test_ss}, the naturalness evaluation task in Subsection \ref{test_naturalness} and the code-switching task in Subsection \ref{test_cs}.

\subsubsection{Speaker Similarity} 
\label{test_ss}

We evaluated 10 trials. Each trial consisted of the same sentence synthesized by all tested TTS models, and the original speaker’s recording was included as one of the stimuli. Participants were also provided with a reference sample of the original speaker’s voice and rated each stimulus based on how similar the voice and intonation were to the reference. This setup allowed us to observe whether participants could correctly identify the original recording among the synthesized samples.

The prompt given for this task was: \texttt{Please rate each audio sample according to how similar the speaker sounds to the reference speaker}.

\begin{figure}[H]

\centering
\includegraphics[width=0.8\textwidth]{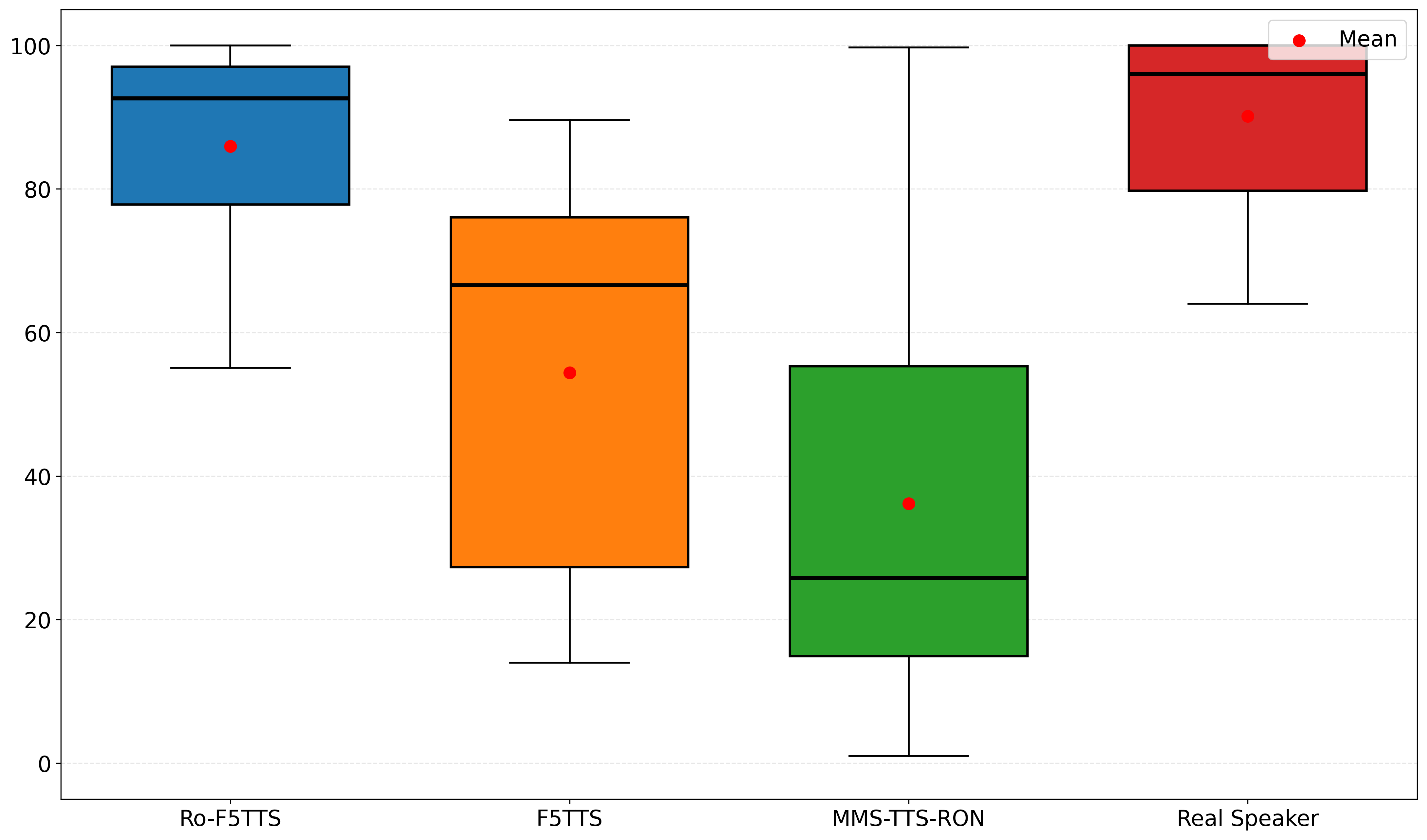}
  \caption{Results on speaker similarity task.}
  \label{fig:similarity}
\end{figure}

We observed that users often found it challenging to correctly identify the reference speaker among the stimuli. The results of each trial are shown in Figure \ref{fig:similarity}. 

We observed that some of the listeners had difficulties identifying the real speaker, with overall high scores for our model, which, even in poor audio settings, still suggests that the model efficiently encompasses speaker identity in generated output, in terms of speaking style and voice timbre.

\subsubsection{Pronunciation and Naturalness} 
\label{test_naturalness}

We tested 9 longer sentences to assess how accurately the speech was pronounced and how natural it sounded. Table \ref{pron-table} displays the sentences synthesized for the task. 

\fontsize{10}{12}{
\begin{table}[H]
\begin{center}
\caption{Sentences used for evaluating pronunciation.}
\label{pron-table} 
\begin{tabular}{|p{12cm}|}
\hline 
\bf Sentence \\ \hline
Am avut astăzi o întâlnire foarte lungă cu clientul, care a durat aproape două ore, și am discutat multe detalii importante despre proiect. \\ \hline
Poți să-mi trimiți și mie, te rog, documentul respectiv, ca să pot verifica toate informațiile înainte de întâlnirea de mâine? \\ \hline
Săptămâna aceasta vreau să mă relaxez acasă, să citesc ceva și poate să vizionez un film sau să mă uit la un serial interesant. \\ \hline
Este extrem de enervant când oamenii nu răspund la mesajele pe care le trimiți, mai ales dacă ai nevoie urgentă de informații. \\ \hline
Îți voi trimite un raport complet mâine dimineață, imediat după ce reușesc să vorbesc cu toți colegii implicați în proiect. \\ \hline
Nu sunt sigur dacă are sens, dar hai să încercăm oricum, chiar dacă există riscul să nu funcționeze perfect. \\ \hline
A spus că termenul limită este vineri, dar sincer, nu cred că va termina tot ce și-a propus până atunci, așa că trebuie să verificăm împreună progresul. \\ \hline
Mă duc să iau o cafea de la cafenea și revin imediat, ca să putem continua discuția fără întreruperi. \\ \hline
Am avut o întâlnire foarte devreme dimineața și nu am avut timp nici măcar să mănânc, așa că am plecat de acasă cu stomacul gol. \\ \hline
\end{tabular}
\end{center}
\end{table}
}

\begin{figure}[H]
\centering
\includegraphics[width=0.8\textwidth]{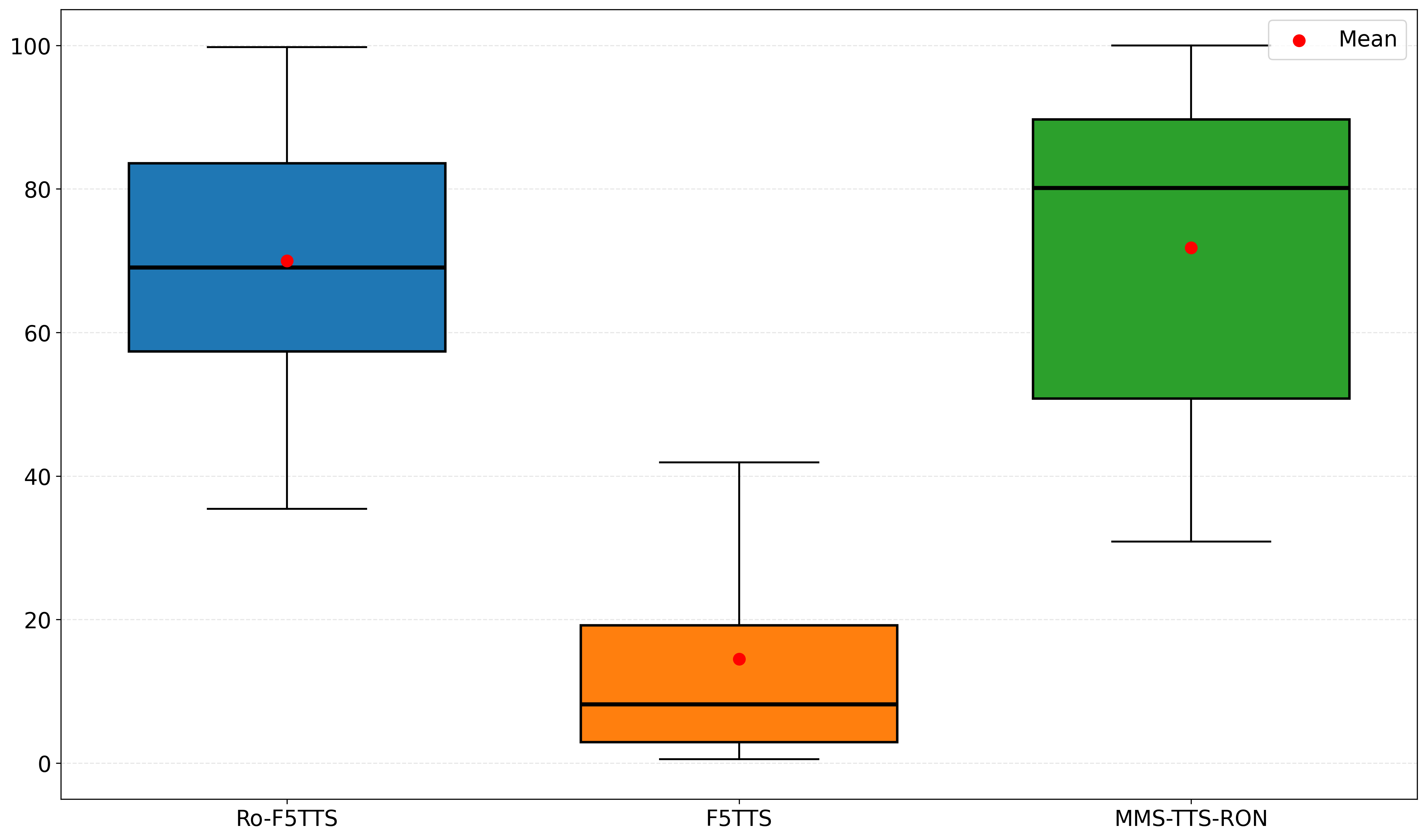}
  \caption{Results on pronunciation and naturalness task.}
  \label{fig:pronuntiation}
\end{figure}

The results are shown in Figure \ref{fig:pronuntiation}.

In this task, the MMS-TTS-RON model achieved higher overall scores for most sentences. Part of this result is due to listeners noticing an English accent in certain examples, which affected their evaluation of our model.

The prompt shown to the listeners was: \texttt{Please rate each audio sample based on the pronunciation of the words and how natural it sounds}.

\subsubsection{Code Switching}
\label{test_cs}

We assessed the model's ability to handle instances where both Romanian and English are used within the same utterance, using 9 examples. Table \ref{sentences-table} displays the code-switching examples included in our evaluation.

\fontsize{10}{12}{
\begin{table}[H]
\begin{center}
\caption{Code-switching sentences used for evaluation.}
\label{sentences-table} 
\begin{tabular}{|p{12cm}|}
\hline 
\bf Sentence \\ \hline
Am avut un call with the client today și a durat două ore. \\ \hline
Poți să-mi trimiți și mie un link to that te rog? \\ \hline
Săptămâna asta vreau să mă relaxez and maybe watch a movie or something. \\ \hline
E super annoying când nu răspunde lumea la mail. \\ \hline
Îți fac un update pe mâine dimineață după ce vorbesc cu ei. \\ \hline
Nu știu dacă are sens dar let's try anyway. \\ \hline
A zis că deadline-ul pe vineri dar actually nu cred că termină. \\ \hline
Mă duc să iau un coffee to go și vin imediat brother. \\ \hline
Am avut un meeting super early și nici nu am apucat să mănânc. \\ \hline
\end{tabular}
\end{center}
\end{table}
}

\begin{figure}[H]
\centering
\includegraphics[width=0.8\textwidth]{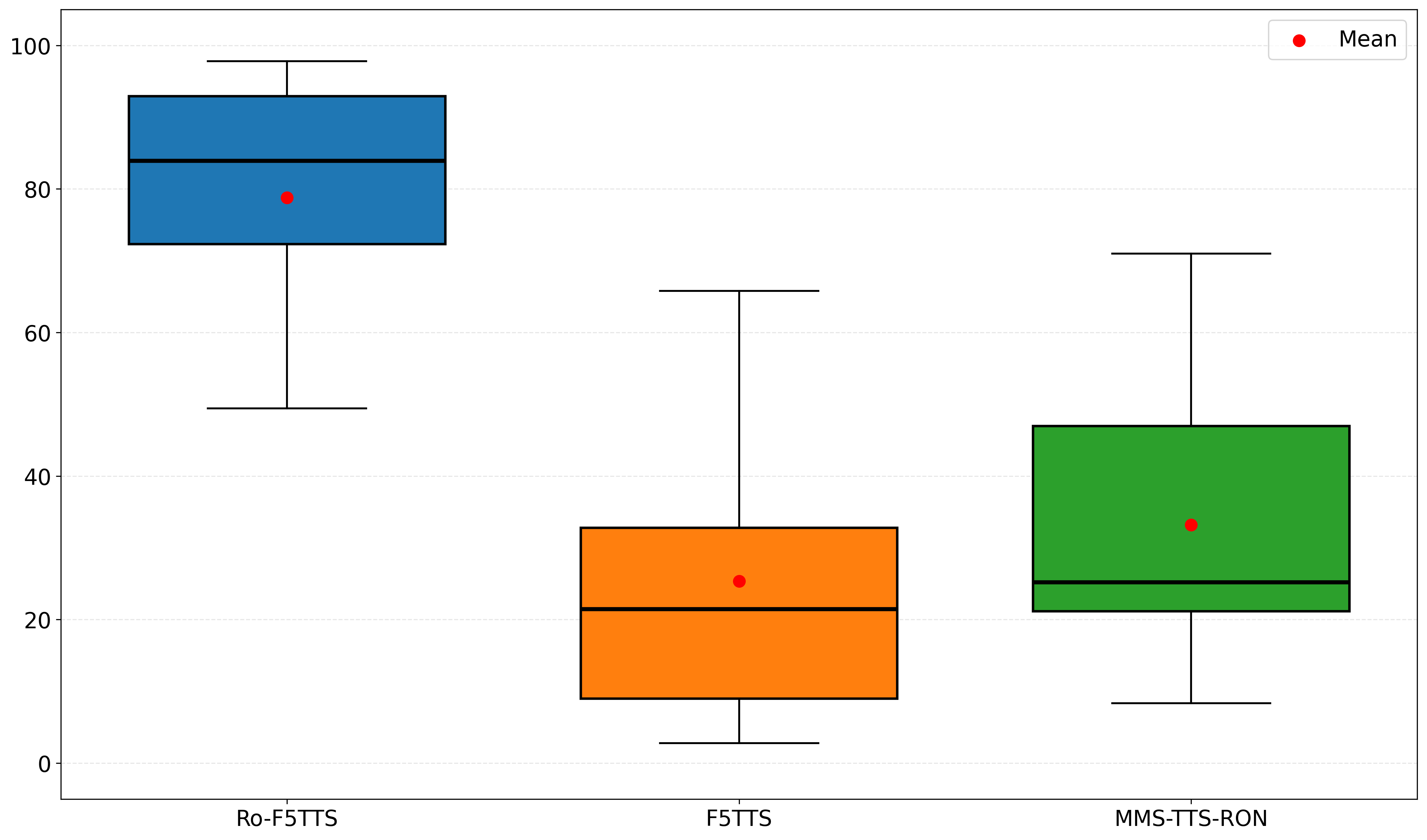}
  \caption{Results on Romanian-English code-switching task.}
  \label{fig:code_switch}
\end{figure}

Results of the evaluation are shown in Figure \ref{fig:code_switch}.

For code-switching, listeners were instructed as follows: \texttt{Please rate each audio sample based on how natural the transition between Romanian and English is}. While the results indicate potential, the tests were limited in size. Further analysis is required, both from a qualitative perspective, examining perceptual naturalness and transition smoothness, and from a mechanistic perspective, to understand what occurs internally during language switching.

The system could benefit from automated language switch detection rather than manual annotation, an integration that we aim to investigate in future work.

\subsection{Objective Evaluation}
\label{evaltion:objective}

For our objective evaluation, we follow the approach from \cite{f5tts} and \cite{eff_fastpitch} and use the transcription and speaker recognition models. In our case, we applied them to a set of \textbf{1000 generated samples} from Common Voice. We also add for comparison results obtained by fully-finetuning the F5-TTS model on our training data.

\subsubsection{Word Error Rate}
The model we selected is gigant/whisper-medium-romanian, a finetuned variant of Whisper \cite{whisper}, available on the Huggingface platform at \cite{gigant_whisper_medium_romanian}. The transcriptions are then used to compute Word Error Rate (WER), Match Error Rate (MER), Word Information Lost (WIL) and Word Information Preserved (WIP) . The results are displayed in Table \ref{metrics-table}.

\fontsize{10}{12}{
\begin{table}[H]
\begin{center}
\caption{Evaluation metrics for Romanian TTS systems.}
\label{metrics-table}
\begin{tabular}{|p{5cm}|p{3cm}|p{3cm}|p{3cm}|}
\hline 
\bf Metric (\%) & \bf MMS-TTS-RON & \bf RO-F5TTS & \bf F5-TTS-FULL-FT \\ \hline
WER (↓) & 5.77 & 5.27 & 3.62 \\ \hline
MER (↓) & 5.63 & 5.15 & 3.52 \\ \hline
WIL (↓) & 9.52 & 8.23 & 5.39 \\ \hline
WIP (↑) & 90.48 & 91.77 & 96.92 \\ 
\hline
\end{tabular}
\end{center}
\end{table}
}

We observe that the fully fine-tuned variant achieved better results, suggesting that the adapter is not fully able to capture all task-specific nuances, particularly the phonetic differences across languages, which likely contributed to lower performance on the pronunciation task.

\subsubsection{Speaker Similarity}
To assess speaker similarity, we employed TitaNet-L to compute speaker embeddings based on audio, following the approach of \cite{eff_fastpitch}.
Cosine similarity was calculated between the reference and generated audio samples.
The results presented in Table \ref{similarity-table} show a consistently high similarity between the reference speaker and the generated voice.

\fontsize{10}{12}{
\begin{table}[H]
\begin{center}
\caption{Cosine Similarity Statistics between reference and generated audio for two models.}
\label{similarity-table} 
\begin{tabular}{|p{5cm}|p{3cm}|p{3cm}|}
\hline 
\bf Statistic & \bf RO-F5TTS & \bf F5-TTS-FULL-FT \\ \hline
Mean & 0.9013 & 0.7946 \\ \hline
Standard Deviation & 0.0319 & 0.0578 \\ \hline
Minimum & 0.7261 & 0.5446 \\ \hline
Maximum & 0.9656 & 0.9186 \\ \hline
Median & 0.9061 & 0.8033 \\ 
\hline
\end{tabular}
\end{center}
\end{table}
}

We observe a substantial improvement over the F5-TTS fine-tuned model. The Speaker Similarity and WER results highlight the trade-off between the current adapter-based approach and full fine-tuning, reflecting a balance between maintaining voice cloning capabilities and accurate language-specific speech synthesis.

\section{Conclusions \& Future Work}
We present a \textbf{small input-level adapter that enables Romanian text synthesis with F5-TTS} while keeping the pretrained model frozen. The adapter replaces the character embedding layer and adds a lightweight 1D ConvNeXt that \textbf{produces continuous ``soft'' letter-to-sound} consumable by the frozen backbone. 

 In subjective tests with 20 native speakers, we observed high speaker similarity. Pronunciation and naturalness were competitive, though occasionally influenced by an English accent. At this stage, it remains unclear whether this limitation arises from the adapter itself, from the underlying base model’s lack of direct training on Romanian, or from the interplay between the two, as the adapter may not be strong enough to represent patterns the model has never encountered.

Objectively, our system achieved a lower WER than MMS-TTS-RON (5.27\% vs.\ 5.77\%) and stronger speaker similarity compared to the fully finetuned F5-TTS.

\textbf{Future work includes}:
\begin{enumerate}[nolistsep]
    \item A detailed analysis of the representations in order to understand how the adapter encodes and projects the textual input with respect to the previously learned embedding space.
    \item Exploring automated language switch detection and explicit bilingual training to improve code-switching performance.
    \item Extending the framework to additional languages and models.
\end{enumerate}

\bibliographystyle{consilr.bst}
\bibliography{main.bib}

\end{document}